%% file: main.tex
\documentclass{article}

\usepackage[table]{xcolor}

\usepackage[preprint]{corl_2025} 

\usepackage{amsmath}
\usepackage{amssymb}
\usepackage{amsfonts}
\usepackage{graphicx}
\usepackage{wrapfig}
\usepackage{caption}
\usepackage[most]{tcolorbox}
\usepackage{booktabs}
\usepackage{enumitem}
\usepackage{threeparttable}
\usepackage{caption}
\usepackage{subcaption} 
\usepackage{listings}

\newenvironment{tightlist}%
{\begin{list}{$\bullet$}{%
    \setlength{\topsep}{0in}
    \setlength{\partopsep}{0in}
    \setlength{\itemsep}{0in}
    \setlength{\parsep}{0in}
    \setlength{\leftmargin}{1.5em}
    \setlength{\rightmargin}{0in}
}
}%
{\end{list}
}

\tcbset{
  base/.style={
    before skip=2em,
    after skip=2em,
    colback=gray!5!white,
    colframe=gray!25!white,
    fonttitle=\bfseries,
    coltitle=black,
    boxrule=0.8pt,
    arc=4pt,
    left=6pt,right=6pt,top=6pt,bottom=6pt
  }
}

\newtcolorbox{examplebox}[1][]{base,title=Example,#1}
\newtcolorbox{resultbox}[1][]{base,title=Result,#1}
\newtcolorbox{emptybox}[1][]{base,title=#1}

\definecolor{codegray}{gray}{0.95}
\title{Coloring Between the Lines: Personalization in the Null Space of Planning Constraints}

\author{
  Tom Silver, Rajat Kumar Jenamani, Ziang Liu, Ben Dodson, Tapomayukh Bhattacharjee\\
  Department of Computer Science\\
  Cornell University
}

\begin{document}
\maketitle
\input{figures/teaser}


\begin{abstract}
\input{abstract}
\end{abstract}


\input{variables}
\input{intro}
\input{related_work}
\input{problem_formulation}
\input{approach}
\input{experiments}
\input{conclusion}

\clearpage

\input{limitations}
\input{acknowledgments}

\bibliography{references}  

\clearpage
\appendix
\input{appendix}

\end{document}

%% file: figures/teaser.tex
\begin{center}
    \captionsetup{type=figure}
    \vspace{-0.3cm}
    \includegraphics[width=\textwidth]{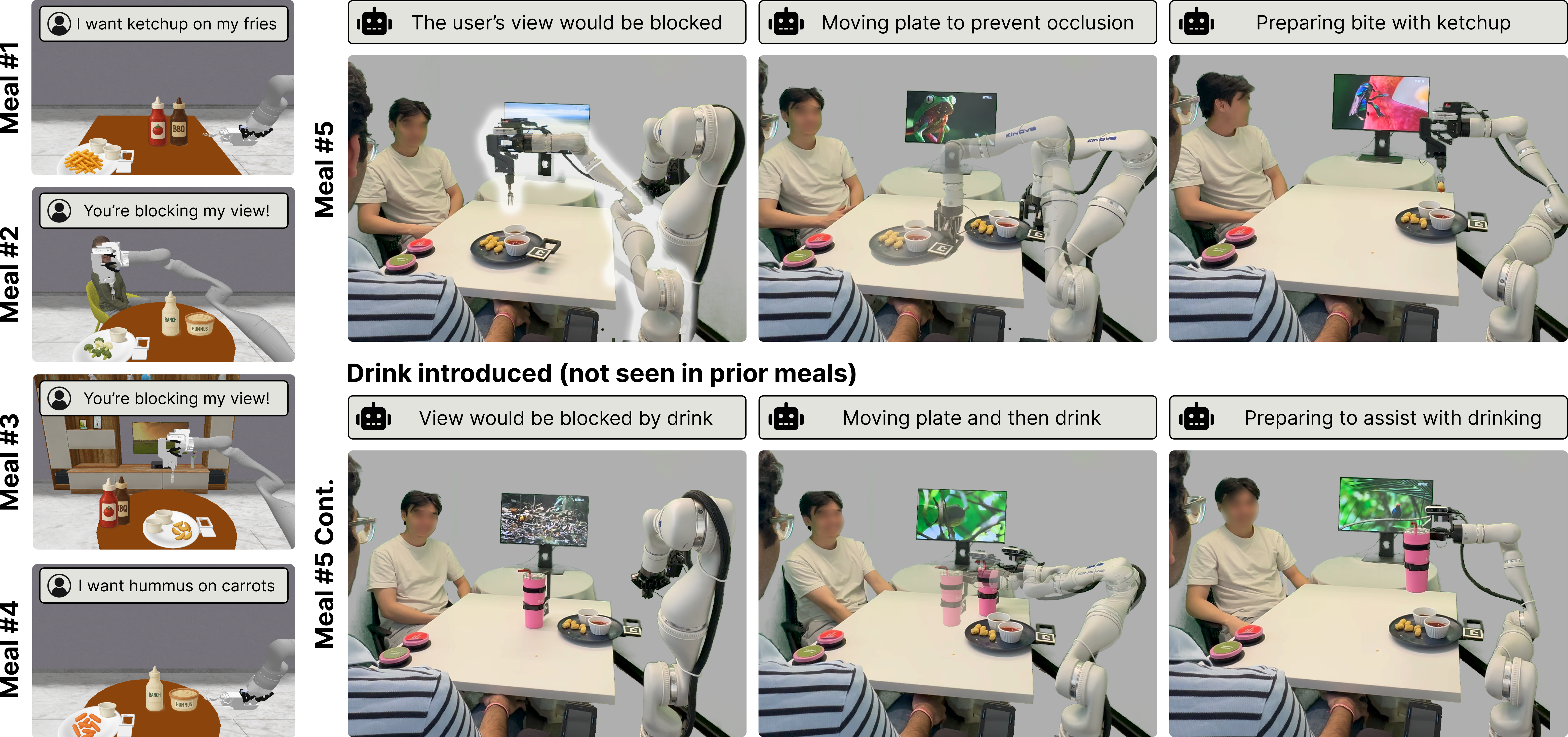}
    \captionof{figure}{
    \textbf{Real robot demonstration of Coloring Between the Lines (CBTL)}. The robot personalizes to the user over five meals (four in sim). During the fifth meal, CBTL plans to reposition the plate to prevent violating the user's learned occlusion preferences. It then chooses ketchup based on learned dipping preferences. Finally, CBTL generalizes to avoid occlusions with a new drink.}
    \label{fig:front}
\end{center}

%% file: abstract.tex
Generalist robots must personalize in-the-wild to meet the diverse needs and preferences of long-term users. How can we enable flexible personalization without sacrificing safety or competency? This paper proposes Coloring Between the Lines (CBTL), a method for personalization that exploits the null space of constraint satisfaction problems (CSPs) used in robot planning. CBTL begins with a CSP generator that ensures safe and competent behavior, then incrementally personalizes behavior by learning parameterized constraints from online interaction. By quantifying uncertainty and leveraging the compositionality of planning constraints, CBTL achieves sample-efficient adaptation without environment resets. We evaluate CBTL in (1) three diverse simulation environments; (2) a web-based user study; and (3) a real-robot assisted feeding system, finding that CBTL consistently achieves more effective personalization with fewer interactions than baselines. Our results demonstrate that CBTL provides a unified and practical approach for continual, flexible, active, and safe robot personalization. Website: \url{https://emprise.cs.cornell.edu/cbtl/}

%% file: variables.tex
\newcommand{\observations}{{\mathcal{O}}}
\newcommand{\observation}{{o}}
\newcommand{\actions}{{\mathcal{A}}}
\newcommand{\action}{{a}}
\newcommand{\history}{{h}}
\newcommand{\horizon}{{T}}
\newcommand{\performance}{{J}}
\newcommand{\cspvariables}{{\mathcal{V}}}
\newcommand{\cspvariable}{{V}}
\newcommand{\cspvalue}{v}
\newcommand{\cspvariablename}{{N(V)}}
\newcommand{\cspvariabledomain}{{D(V)}}
\newcommand{\cspconstraints}{{\mathcal{C}}}
\newcommand{\cspconstraint}{{C}}
\newcommand{\cspconstraintfunction}{{c}}
\newcommand{\cspsolution}{{\mathbf{v^*}}}
\newcommand{\cspcandidatesolution}{{\mathbf{v}}}
\newcommand{\cspgenerator}{\textsc{CSPGen}}
\newcommand{\cspsampler}{\psi}
\newcommand{\policy}{\pi}
\newcommand{\termination}{\tau}
\newcommand{\personalizedconstraints}{\mathcal{C}_p}
\newcommand{\personalizedconstraint}{C_p}
\newcommand{\personalizedconstraintgen}{\textsc{Gen}_p}
\newcommand{\initiation}{\iota_p}
\newcommand{\constraintparameters}{\theta}
\newcommand{\classificationlabel}{\ell}
\newcommand{\naturallanguagelabel}{w}
\newcommand{\entropy}{\mathcal{H}}
\newcommand{\dataset}{\mathcal{D}}
\newcommand{\numparticipants}{60}

%% file: intro.tex
\section{Introduction}
\label{sec:intro}

When a generalist robot first arrives at a new home, restaurant, or hospital, it should already know a great deal---how to move, how to manipulate objects, and how to interpret human-given commands, among many other general competencies.
But the robot will not yet know about the aspects of its environment that make it unique.
It will not yet know that a user loves tacos but hates cilantro; that they have limited range of motion~\cite{liu2025grace}; or that they sometimes like to watch TV while eating dinner.
The robot should learn these nuances over time and \emph{personalize} its behavior accordingly~\cite{liu2020learning,madan2022sparcs,patel2024robot,leap-hri-2025}.

Continual personalization of a generalist robot has been a longstanding goal in robotics, but it has remained elusive because of the inherent tension between \emph{safety} and \emph{flexibility}.
An over-constrained robot cannot adapt; an under-constrained robot can be dangerous.
Previous work on personalization has largely deferred this challenge by focusing on task-specific forms of personalization~\cite{gao2015user,saunders2015teach,canal2016personalization,zhang2017personalized,wilde2018learning,canal2019adapting,wu2023tidybot,jenamani2025feast} or by working in settings without critical safety requirements~\cite{wei2024towards,chen2024large,zheng2025lifelong}.

In this work, we propose a general method for resolving these competing concerns.
We start by leveraging \emph{planning with compositional constraints}~\cite{lozano2014constraint,garrett2021integrated,curtis2024trust,thummtext2interaction,huang2024rekep} as a framework to enable a broad range of behavior while adhering to explicit programmer-specified restrictions~\cite{wachi2024survey}.
For example, consider the mealtime assistance system shown in Figure~\ref{fig:front}.
Given a request for a bite of food, the robot uses its given non-personalized capabilities to generate a constraint satisfaction problem (CSP) with variables $\mathrm{platePose} \in \text{SE(2)}$, $\mathrm{utensilAbovePlatePose} \in \text{SE(3)}$, $\mathrm{robotAbovePlateConf} \in \mathbb{R}^7$, $\mathrm{dip} \in \{\text{ketchup}, \text{BBQ sauce}, \text{none}\}$, and $\mathrm{readySignal} \in \{\text{mouth open}, \text{button}, \text{auto}\}$, among others, which are mutually constrained, e.g., to enforce kinematic feasibility and safe collision-free motion.

CSPs generally have multiple solutions, especially when continuous variables are involved.
We refer to the set of solutions to a CSP as a \emph{null space} by analogy to the concept in kinematics~\cite{nakamura1987task,khatib1987unified,sciavicco2001modelling} which refers to the fact that a redundant-DoF manipulator can move without affecting the end-effector's task, e.g., repositioning the elbow while keeping the hand fixed.
Our key insight is that \emph{this CSP null space provides an opportunity for personalization.}
Rather than picking arbitrary solutions, the robot should ``color between the lines'' of the constraints to select solutions in the null space that are personalized to the user.
In particular, we propose that robot engineers should not only implement one-size-fits-all constraints to ensure safety and general competency---they should also implement constraints with learnable parameters that can be rapidly personalized in-the-wild.

For example, previous work in assisted feeding has shown that users vary in their bite preferences (e.g., dipping sauce)~\cite{jenamani2024flair}, their interaction modes (e.g., ready signals)~\cite{jenamani2025feast}, and their sensitivities to visual occlusions (e.g., by the robot arm)~\cite{belkhale2022balancing}.
Our approach unifies these and other aspects of personalization that previous work has treated task-dependently.
For instance, we generate a $\mathrm{dipPreferred}(\mathrm{dip}, \constraintparameters_1)$ constraint where $\constraintparameters_1 \in \mathrm{Text}$ is a natural language summary of the user's dip preferences, and the constraint itself is implemented with a large language model (LLM)~\cite{openai2024gpt4omini}; a similar $\mathrm{signalPreferred}(\mathrm{readySignal}, \constraintparameters_2)$ constraint where $\constraintparameters_2 \in \mathrm{Text}$; and an $\mathrm{occlusionFree}(\mathrm{conf}, \constraintparameters_3)$ constraint, where $\constraintparameters_3 \in [0, 1]$ encodes the user's occlusion sensitivity and where the constraint itself uses a model from prior work~\cite{belkhale2022balancing}.
As the robot gathers experience, it adapts the parameters $\constraintparameters_1, \constraintparameters_2, \constraintparameters_3$ and continually personalizes its behavior.

To maintain flexibility, it is essential that the robot learns full personalized constraints and does not simply memorize CSP solutions.
For example, plate and drink positions that satisfied the user in the past may become impossible if the table changes or if obstacles are introduced.
We therefore propose to \emph{actively learn} personalized constraint parameters using an entropy-based method~\cite{settles2009active,li2023embodied}.
This enables the robot to rapidly learn personalized constraints that generalize across scenarios.

We call this overall approach---planning with compositional constraints, learning personalized constraint parameters, and actively gathering data---\emph{Coloring Between the Lines (CBTL)}.
In experiments, we evaluate the extent to which CBTL achieves efficient and effective personalization.
We first compare CBTL to four baselines in three simulation environments and find that CBTL consistently personalizes better and faster.
We then conduct a web-based user study with \numparticipants{} participants and find that they significantly prefer CBTL's choices over a non-personalized baseline.
Finally, we validate the method with the real robot in Figure~\ref{fig:front} and show that CBTL generalizes: by reusing the personalized occlusion constraint learned from feeding, CBTL anticipates new occlusions and repositions the drink accordingly.
Furthermore, to avoid collisions and maintain reachability, CBTL repositions the plate to make room for the drink.
Together, these results demonstrate that CBTL provides a unified and practical approach for continual, flexible, active, and safe robot personalization.

%% file: related_work.tex
\section{Related Work}
\label{sec:related-work}

\textbf{Personalization in Robot Learning. }
In this work, we develop a method for robot personalization that is \emph{continual}, \emph{flexible}, \emph{active}, and \emph{safe}.
Previous work considers one-time personalization in the context of single tasks~\cite{gao2015user,saunders2015teach,canal2016personalization,zhang2017personalized,wilde2018learning,canal2019adapting,belkhale2022balancing,brawer2023interactive}.
For example, given natural-language user preferences such as ``feed me spaghetti before meatballs''~\cite{jenamani2024flair} or ``sort the laundry by lights and darks''~\cite{wu2023tidybot}, robot behavior can be adapted on-the-fly to take these preferences into account.
Some of these works explicitly consider safety~\cite{thummtext2interaction,karagulle2024safe,mukherjee2024personalization}.
The works of \citet{thummtext2interaction} and \citet{wang2024apricot} are especially relevant; like us, they consider personalization within constraint-based systems and focus respectively on safety and active learning.
However, all of these approaches require users to fully specify their preferences at the beginning of every task; the robot does not learn online over time or generalize between tasks.
Closer to our goal are efforts to learn user models that can be reused across multiple tasks~\cite{hellou2021personalization,he2023learning,wang2024personalization,liu2025grace}.
These works primarily address the user models themselves and do not consider continual or active learning.
Separate work on lifelong personalization~\cite{lee2012personalization,leyzberg2018effect,churamani2020continual,spaulding2021lifelong} primarily considers social elements of human-robot interaction, rather than algorithmic frameworks for achieving personalization.
To the best of our knowledge, CBTL is the first method for robot personalization that is continual, flexible, active, and safe.

\textbf{Learning in Null Space. } Our key insight in this work is that planning with constraints induces a ``null space'' that enables safe personalization. 
Other work in robot learning has considered learning in null space, typically in the more literal case where the null space refers to the redundant degrees of freedom of a manipulator~\cite{d2001learning,salaun2009control,towell2010learning,nordmann2012teaching,yang2021null}.
There is also work in safe reinforcement learning that generalizes the concept of null spaces to constraint manifolds and focuses on learning new low-level policies~\cite{liu2022robot,liu2024safe}.
These works do not consider personalization and typically focus on single tasks.

\textbf{Learning Constraints for Planning. }
Our constraint generation framework takes inspiration from task and motion planning (TAMP)~\cite{garrett2021integrated}.
Learning constraints for TAMP has been previously considered~\cite{mishra2023generative,lin2023text2motion,huang2024rekep,curtis2024trust,loula2020learning}.
Learning predicates for TAMP is also closely related~\cite{silver2023predicate,li2025bilevel,liang2024visualpredicator,athalye2024predicate}.
Especially relevant are the works of \citet{wang2021learning} and \citet{li2023embodied}, who consider \emph{active} constraint learning.
Our use of entropy-based active learning is inspired by the latter work.
These works do not consider personalization.
Other recent work in TAMP has considered active  learning for samplers instead of constraints~\cite{kumar2024practice,mendez2023embodied}.
In principle, these efforts could be combined with our approach to develop a robot that not only personalizes, but also accelerates its planning over time.

%% file: problem_formulation.tex
\section{Problem Formulation}
\label{sec:problem-formulation}
We consider a robot that takes actions $\action \in \actions$ and receives observations $\observation \in \observations$ in an environment that is governed by unknown transition and observation models.
The environment is never reset---the robot has a single trajectory of experience.
Let $\history_t = (o_0, a_1, o_1, \dots, a_t, o_t)$ denote the history of observations and actions through time $t$.
In experiments, we measure the performance of the robot  according to a domain-specific objective over histories $\performance(\history_t) \in \mathbb{R}$.
For example, in assistive feeding, we use a 5-point Likert scale to measure overall user satisfaction.
Unlike in reinforcement learning and related paradigms, we do not assume that the robot can observe $\performance(\history_t)$; the objective is only used to externally evaluate the robot's behavior in experiments.

%% file: approach.tex
\begin{figure}[t]
    \centering
    \includegraphics[width=\linewidth]{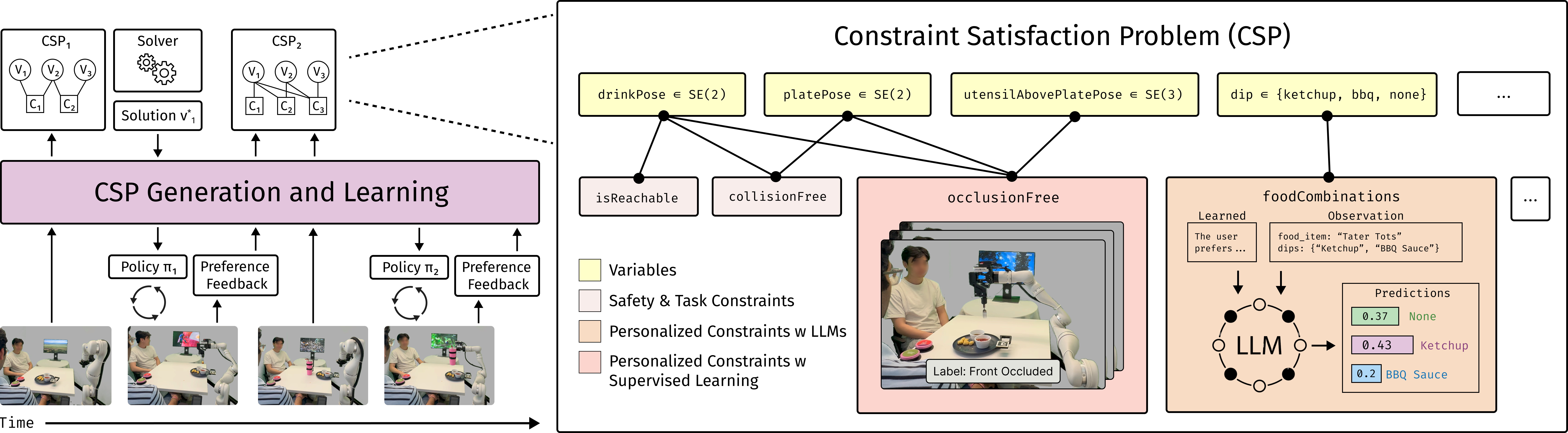}
    \vspace{-1em}
    \caption{\textbf{Overview of Coloring Between the Lines (CBTL)}, our method for robot personalization.}
    \label{fig:cbtl-overview}
    \vspace{-1.5em}
\end{figure}

\section{Personalizing Robots by Coloring Between the Lines}
\label{sec:approach}

We now present Coloring Between the Lines (CBTL), our approach for continual, flexible, active, and safe robot personalization.
See Figure~\ref{fig:cbtl-overview} for an overview of the approach.

\textbf{Decision-Making with CSPs. }
CBTL proposes that robots in-the-wild should make decisions by repeatedly generating and solving constraint satisfaction problems (CSPs).
A CSP $(\cspvariables, \cspconstraints)$ is defined by variables $\cspvariable \in \cspvariables$ and constraints $\cspconstraint \in \cspconstraints$.
Each variable $\cspvariable$ has a unique name $\cspvariablename$ and a domain $\cspvariabledomain$ that may be, for example, a real-valued vector space, a finite set, the set of all natural language strings, or a space of object poses.
A constraint $\cspconstraint$ is defined over one or more variables $\cspvariable_1, \dots, \cspvariable_k$ via a function
$\cspconstraintfunction : D(V_1) \times \cdots \times D(V_k) \to \{\text{True}, \text{False}\}.$
A CSP solution $\cspsolution$ is a joint assignment of all variables to values that satisfies all constraints.
There are many techniques for solving CSPs.
In this work, we use a simple and flexible sampling-based solver (see Appendix~\ref{app:csp-details}).

\vspace{-1em}
\begin{examplebox}
In assisted feeding (Figure~\ref{fig:front}), where the robot can move the drink and plate on the table before attaching a utensil to its end effector, CSP variable examples include $\mathrm{drinkPose} \in \textrm{SE(2)}$, $\mathrm{platePose} \in \textrm{SE(2)}$, $\mathrm{utensilAbovePlatePose} \in \textrm{SE(3)}$, and $\mathrm{robotAbovePlateConf} \in \mathbb{R}^7$, among others.\footnotemark\
Constraint examples include $\mathrm{collisionFree}(\mathrm{drinkPose}, \mathrm{platePose})$, $\mathrm{poseIsAbove}(\mathrm{platePose}, \mathrm{utensilAbovePlatePose})$, and $\mathrm{kinematicallyValid}(\mathrm{utensilAbovePlatePose}, \mathrm{robotAbovePlateConf})$.
\end{examplebox}
\vspace{-1em}
\footnotetext{The notation $\mathrm{platePose} \in \textrm{SE(2)}$ is shorthand for $\cspvariable$ with $\cspvariablename = \mathrm{platePose}$ and $\cspvariabledomain = \textrm{SE(2)}$.
(Note that we use $\textrm{SE(2)}$ because the plate is radially asymmetric given the handle and food.)
Similarly, $\mathrm{collisionFree}(\mathrm{platePose}, \mathrm{drinkPose})$ is shorthand for $\cspconstraint$ over $\mathrm{platePose}$ and $\mathrm{drinkPose}$.
}

\textbf{Generating CSPs. }
We propose that robots should be equipped with  \emph{CSP generators}.
A CSP generator $\cspgenerator$ takes a history $\history_t$ as input and produces (1) a CSP, (2) a solution-conditioned policy $\policy(\action_t \mid \history_{t}, \cspsolution)$, and (3) a solution-conditioned termination condition $\termination(\history_{t}, \cspsolution)$.
After finding a solution $\cspsolution$ to the CSP, the robot takes actions following the policy until the termination condition is satisfied, at which point the next CSP is generated.
If no CSP solution is found, a default fallback policy is used.
This work is agnostic to the implementation of the CSP generator, but
in practice, we take inspiration from TAMP methods like PDDLStream~\cite{garrett2020pddlstream}.

\vspace{-1em}
\begin{examplebox}
When the robot observes in $\history_t$ that the user is requesting a bite, it generates a bite-specific CSP that includes the variables and constraints mentioned above, among others.
After solving the CSP, the robot generates a policy $\policy$ over low-level robot joint commands that starts by moving the drink and plate to the poses found in $\cspsolution$.
The termination condition $\tau$ checks for the successful transfer of a bite, or for a new request (e.g., drinking).
\end{examplebox}
\vspace{-1em}

The CSPs generated by the robot will typically have many solutions.
Our key insight in this work is that this \emph{null space} provides an opportunity for personalization.
The robot should ``color between the lines'' by generating and adapting personalized constraints that further reduce the null space.

\textbf{Generating Personalized Constraints. }
Given a CSP $(\cspvariables, \cspconstraints)$ produced by $\cspgenerator$ from history $\history_t$, CBTL generates additional \emph{personalized constraints} $\personalizedconstraints$ to combine with the original constraints $\cspconstraints$.
The resulting CSP $(\cspvariables, \cspconstraints \cup \personalizedconstraints)$ is solved and the solution is used to condition the policy and termination condition as before.
Personalized constraints $\personalizedconstraint \in \personalizedconstraints$ are produced by a collection of generators $\personalizedconstraintgen^{1}, \dots, \personalizedconstraintgen^{n}$ to enable compositional generalization.
Each generator $\personalizedconstraintgen^i$ takes the original CSP $(\cspvariables, \cspconstraints)$ and the history $\history_t$ as input and returns a (possibly empty) set of personalized constraints $\personalizedconstraints^i$.
The final set of personalized constraints is the union $\personalizedconstraints = \bigcup_{i=1}^n \personalizedconstraints^i$.

Each generator $\personalizedconstraintgen$ is characterized by an \emph{arity} $m$, an \emph{initiation condition} $\initiation$, and a \emph{parameterized constraint function} $\cspconstraintfunction_\constraintparameters$.
The arity refers to the number of variables that the constraint function expects as input.
The initiation condition takes in a tuple of $m$ variables from $\cspvariables$ and determines based on their names and domains whether or not a constraint should be produced for those variables.
The function $\cspconstraintfunction_\constraintparameters$ is used to define the constraints.
The parameters $\constraintparameters$ are constant during generation but updated through learning (see below).
All together, the generator outputs:
\begin{equation*}
\{ ((\cspvariable_1, \dots, \cspvariable_m), \cspconstraintfunction_\constraintparameters) : \initiation(\cspvariable_1, \dots, \cspvariable_m) = \text{True}, \cspvariable_i \in \cspvariables \}.
\vspace{-1em}
\end{equation*}
\vspace{-1.5em}
\begin{examplebox}
The generator for the personalized constraint $\mathrm{occlusionFree}$ has arity $m=1$.
The initiation condition $\initiation$ evaluates to true on variable $\cspvariable$ if $\cspvariablename$ indicates that the variable represents some robot joint configuration.
The constraint $\cspconstraintfunction_{\constraintparameters_{\text{occ}}}$ simulates the robot joint configuration and checks if an occlusion score (computed via~\cite{belkhale2022balancing}) exceeds the learned personalized threshold $\constraintparameters_{\text{occ}}$.
\end{examplebox}
\vspace{-1em}

\textbf{Learning Personalized Constraints. }
CBTL uses the history $\history_t$ collected by the robot to update personalized constraint parameters $\constraintparameters$ for each generator $\personalizedconstraintgen$.
We consider two types of learning.

\emph{Supervised Classification. }
Our first approach uses the history of experience to derive a classification dataset $\dataset$ with elements of the form $(\cspcandidatesolution, \classificationlabel)$, where $\cspcandidatesolution$ is an input to the classifier $\cspconstraintfunction_\constraintparameters$ and $\ell$ is True or False.
The logic for deriving these inputs and outputs varies per constraint.
Given the (initially empty) dataset, standard techniques can be used to optimize the parameters $\constraintparameters$ (see Appendix~\ref{app:csp-details}).

\vspace{-1em}
\begin{examplebox}
The user is sometimes asked whether their view is blocked.
This question and their response are included in $\observation$.
When present, the robot extracts the response and its current joint configuration from $\observation$ and adds the tuple to its classification dataset for the $\mathrm{occlusionFree}$ constraint.
\end{examplebox}
\vspace{-1em}

\emph{Learning with Natural Language. }
Our second approach similarly uses history to derive a dataset $\dataset$, but with elements of the form $(\cspcandidatesolution, \observation, \naturallanguagelabel)$ where $\observation$ is an observation and $\naturallanguagelabel$ is a natural-language string.
These data are used to learn constraints that are implemented with large language models (LLMs).
In particular, we consider constraints of the form $$\cspconstraintfunction_\constraintparameters(\cspvalue_1, \dots, \cspvalue_m) = \textsc{QueryLLM}(\textsc{ConstraintPrompt}(\cspvalue_1, \dots, \cspvalue_m, \observation, \constraintparameters))$$
where $\theta$ is a natural-language string and $\textsc{ConstraintPrompt}$ is a constraint-specific function.
A second constraint-specific function $\textsc{LearningPrompt}$ is used to update the parameter constraints:
$$\constraintparameters \gets \textsc{QueryLLM}(\textsc{LearningPrompt}(\dataset)).$$
See Appendix~\ref{app:llm-details} for LLM constraint generation and learning details.

\vspace{-1em}
\begin{examplebox}
The user is sometimes asked which dipping sauce (if any) they would like for their next bite given the food items in the observation $\observation$.
Their response $\naturallanguagelabel$ and the food items are available in $\dataset$.
To learn the $\mathrm{dipPreferred}(\mathrm{dip}, \constraintparameters)$ constraint, a $\textsc{LearningPrompt}(\dataset)$ collates all food items and responses and asks an LLM to generate a summary of dipping preferences $\constraintparameters$.
A $\textsc{ConstraintPrompt}(\cspvalue, \observation, \constraintparameters) $ extracts the current food items from $\observation$ and asks an LLM whether a user with preferences $\constraintparameters$ would enjoy the dip $\cspvalue$.
\end{examplebox}
\vspace{-1em}

\textbf{Active Learning for Data-Efficient Personalization. }
Just as the null space enables personalization, it also enables exploration---the robot should choose actions within the null space that lead to efficient learning.
We propose a maximum-entropy-based approach~\cite{settles2009active,li2023embodied}.    
When the robot is permitted to explore, rather than treating $\personalizedconstraints$ as personalized constraints that need to be satisfied, we create a mathematical program
\[
\max_{\cspcandidatesolution} \frac{1}{|\personalizedconstraints|}\sum_{\personalizedconstraint \in \personalizedconstraints}\entropy(\personalizedconstraint(\cspcandidatesolution))
\]
\[
\text{subject to } \cspconstraint(\cspcandidatesolution) = \text{True } \forall \cspconstraint \in \cspconstraints
\]
where $\entropy$ is the entropy of the Bernoulli distribution $P(\personalizedconstraint(\cspcandidatesolution) = \text{True})$.
Thus, the robot will choose CSP solutions that maximize its uncertainty about the personalized constraint parameters that it is learning, while still satisfying all non-personalized constraints to guarantee safety and competency.

\vspace{-1em}
\begin{examplebox}
\begin{wrapfigure}{r}{0.5\textwidth}
  \centering
  \vspace{-1em}
  \includegraphics[width=0.5\textwidth]{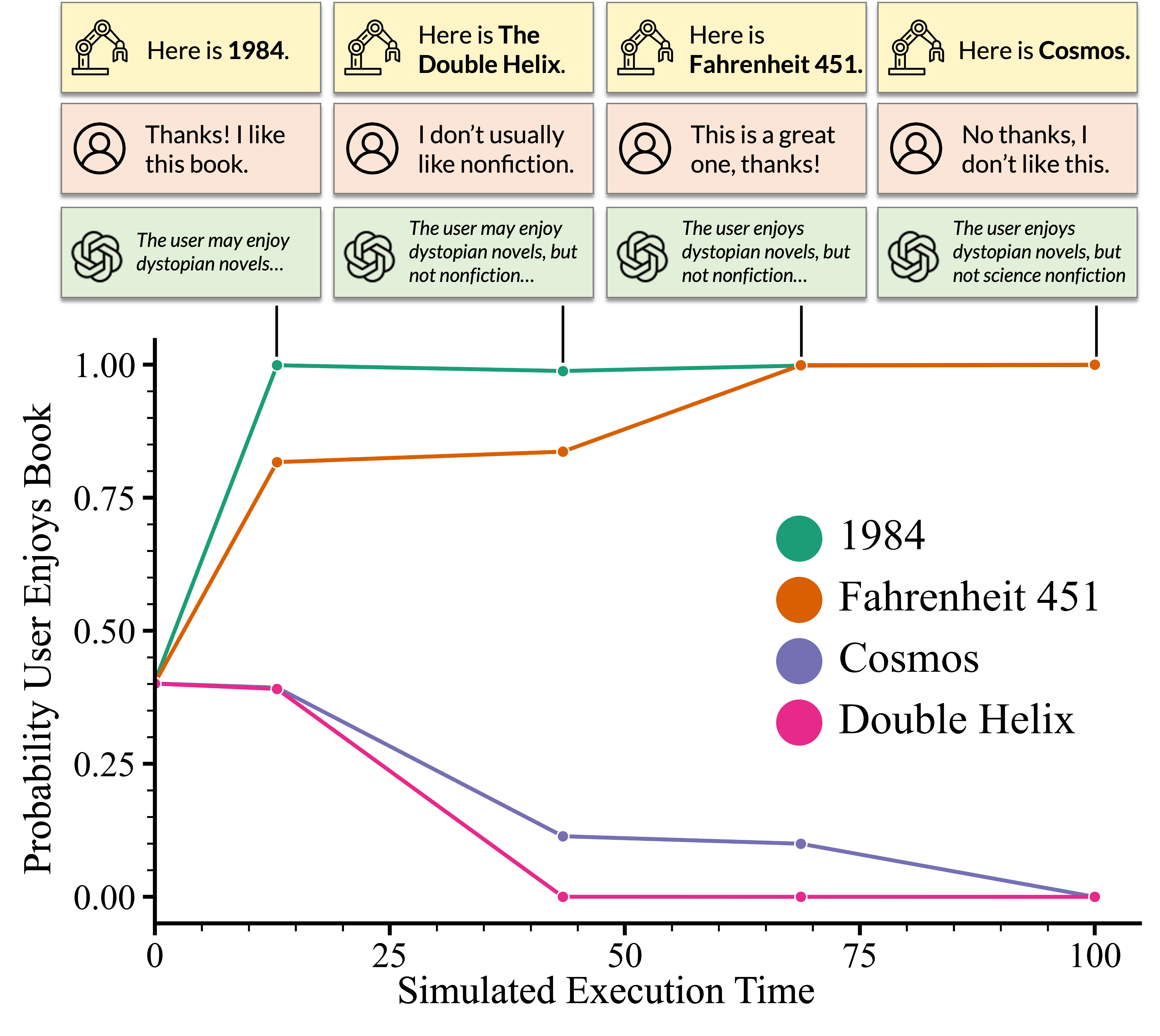}
  \vspace{-1em}
  \caption{CBTL actively reduces uncertainty and generalizes from feedback.}
  \label{fig:books-example}
  \vspace{-1em}
\end{wrapfigure}

For this example, we consider the Books simulation environment (Figure~\ref{fig:books-example}), where a robot must select, retrieve, and hand over books.
CBTL actively learns an LLM-based $\mathrm{userEnjoysBook}(\mathrm{book}, \constraintparameters)$ constraint where $\mathrm{book}$ is a well-known title and author and $\constraintparameters$ is a natural language summary of the user's book preferences.
At first, the robot does not have any information about the user, but after receiving positive feedback about the book ``1984'', it infers that they may also enjoy ``Fahrenheit 451'', another dystopian novel.
To maximize entropy, the robot selects ``The Double Helix'', and learns that the user does not enjoy nonfiction, which implies that they would not like ``Cosmos.''
While learning book preferences, CBTL also actively selects handover poses to rapidly learn the user's functional range of motion~\cite{liu2025grace} (Section~\ref{sec:experiments}).

\end{examplebox}
\vspace{-1.5em}

%% file: experiments.tex
\section{Experiments and Results}
\label{sec:experiments}

In this section, we present simulation experiments, a web-based user study with \numparticipants{} participants, and a real-robot demonstration with a mealtime assistance system.
In addition to assessing the overall \emph{effectiveness}, \emph{efficiency}, and \emph{generality} of CBTL, we consider the following questions:
\begin{tightlist}
    \item[1.] To what extent is \emph{active learning} important for sample-efficient personalization?
    \item[2.] To what extent is CBTL able to personalize to \emph{real human preferences}?
    \item[3.] To what extent is CBTL able to \emph{rapidly generalize} during continual personalization?
\end{tightlist}
\vspace{-0.5em}

\begin{figure}
    \centering
    \includegraphics[width=0.91\linewidth]{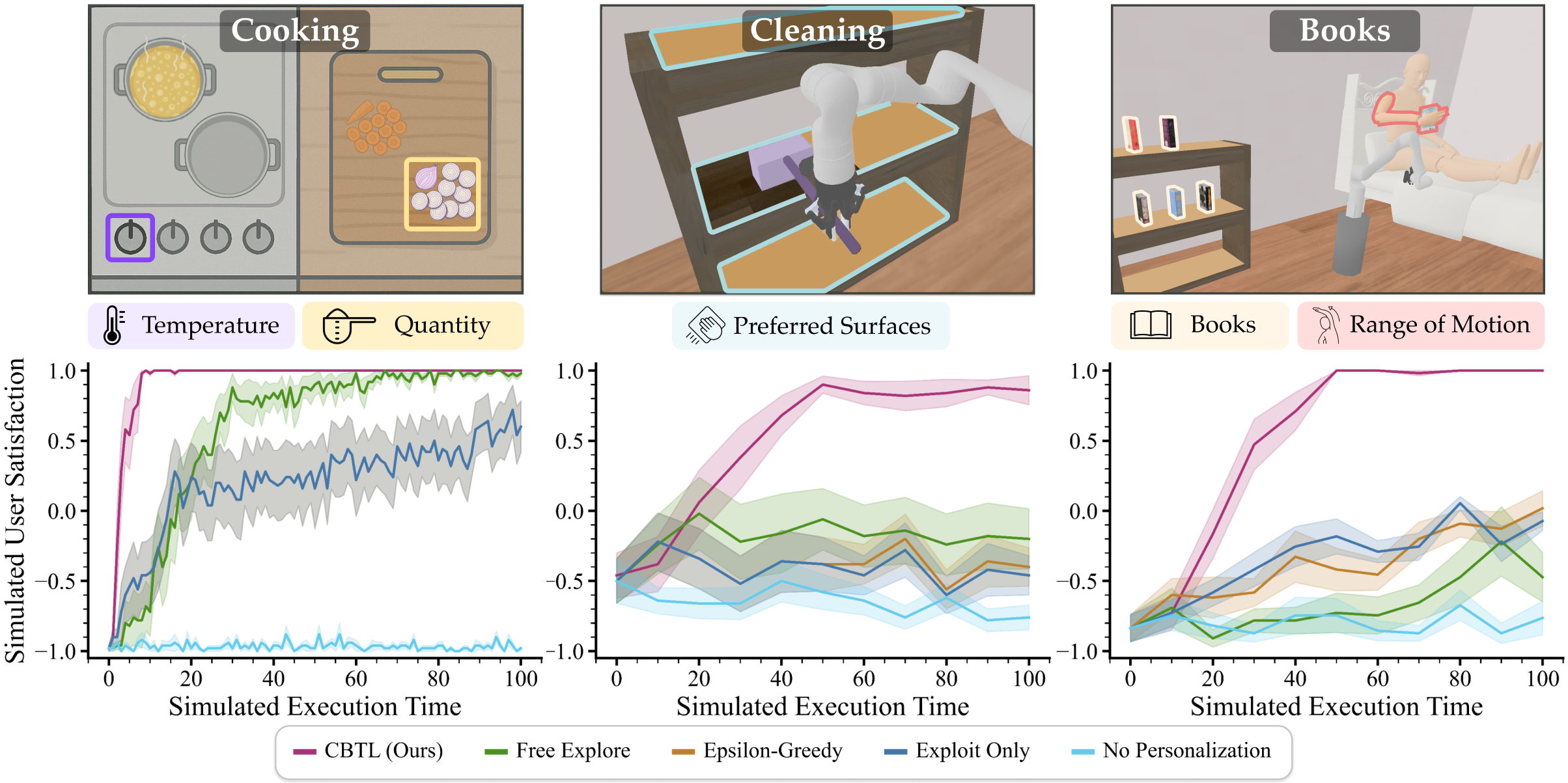}
    \caption{\textbf{Main simulation results.} Lines are means and shaded regions are standard errors over 10 seeds. Units are deliberately omitted (time is arbitrary in simulation and user satisfaction is environment-specific). In Cooking, Exploit Only and Epsilon-Greedy obtain equal performance; see Appendix~\ref{app:sim-experiment-details} for discussion. CBTL (ours) consistently personalizes better and faster than baselines.}
    \label{fig:main-sim-results}
    \vspace{-1.5em}
\end{figure}

\subsection{Simulation Experiments}

We first compare CBTL to four baselines in three diverse simulation environments.
All baselines are given the same prior knowledge in the form of CSP generators.
For each approach in each environment, we run 10 random seeds.
We evaluate performance by loading checkpoints and running held-out tasks with simulated user satisfaction metrics.
We now briefly describe the environments and baselines with details in Appendix~\ref{app:sim-experiment-details}.
We also release code for reproducing all results.

\textbf{Environments. }
We begin with a brief description of the simulation environments.

\begin{tightlist}
    \item \textbf{Cooking}: a 2D stove-top environment that features \emph{pots} with continuous positions and sizes; \emph{ingredients} with continuous quantities and temperatures; a \emph{user} with unknown  preferences. The robot must move pots onto the stove, put ingredients in the pot, wait for them to heat, and then serve the dish to the user. When served a meal, the user gives feedback about ingredient quantity and temperature. CSP non-personalized constraints include: pots cannot overlap on the stove; ingredients can only be used in the quantities available; ingredients must comprise some known meal. A learned personalized constraint enforces that the user enjoys the served meal.
    \item \textbf{Cleaning}: a PyBullet~\cite{coumans2016pybullet} environment where a Kinova Gen3 7-DoF arm~\cite{kinova} on a holonomic mobile base must use a dusting tool to clean tables and shelves. A user watching the robot has unknown preferences about which surfaces should be cleaned and admonishes the robot if it violates their preferences. CSP non-personalized constraints enforce  kinematic feasibility and collision-free motion for moving the robot base, grasping the dusting tool, and positioning the tool to wipe. A learned personalized constraint enforces that only preferred surfaces are cleaned.
    \item \textbf{Books}: a PyBullet~\cite{coumans2016pybullet} environment where a Kinova Gen3 7-DoF arm~\cite{kinova} on a holonomic mobile base must fetch books from a shelf and hand them over to a user in bed.
    The user has unknown book preferences and an unknown functional range of motion~\cite{liu2025grace}, outside of which handovers fail. CSP non-personalized constraints enforce  kinematic feasibility and collision-free motion for moving the robot base, grasping and placing books, and handing over books. Learned personalized constraints enforce reachable handovers and book choices the user prefers.
\end{tightlist}

\textbf{Baselines. } We now briefly describe the baselines used in the simulation experiments.
\begin{tightlist}
    \item \textbf{No Personalization}: A naive baseline where personalized constraints are never updated.
    \item \textbf{Free Explore}: A baseline that explores by fully excluding all personalized constraints during CSP generation. Personalized constraints are still learned and used during evaluation.
    \item \textbf{Exploit Only}: A baseline that always enforces the learned personalized constraints during both exploration and evaluation.
    \item \textbf{Epsilon-Greedy}: A baseline that uses Free Explore $\epsilon$-percent of the time and Exploit Only otherwise, similar to epsilon-greedy exploration in reinforcement learning.
\end{tightlist}

\textbf{Results and Analysis. }
In Figure~\ref{fig:main-sim-results}, we find that CBTL consistently personalizes better and faster than baselines across all three environments.
For example, in Books, the Exploit Only baseline overfits to the first book and handover pose that satisfy the user, failing to learn about the user's full preferences and range of motion.
Free Explore does not overfit, but its exploration is unguided, leading the robot to waste time.
Epsilon-Greedy balances exploration and exploitation, but it is also unguided.
Overall, these findings confirm the importance of entropy-based exploration in CBTL.

\subsection{Web-Based User Study}

We next conduct a web-based user study to evaluate the extent to which CBTL can personalize to real human preferences.
We briefly describe the study and give additional details in Appendix~\ref{app:web-study}.

\begin{figure}
    \centering
    \includegraphics[width=\linewidth]{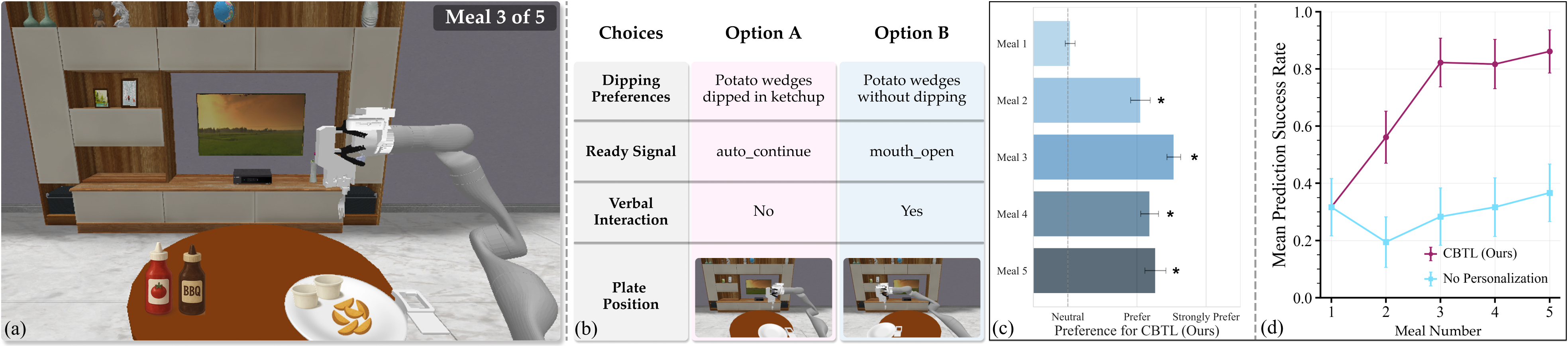}
    \caption{\textbf{Web-based user study.} (a) Example meal. (b) The user is asked which set of choices they prefer on a 5-point Likert scale. (c) Users significantly ($p < 0.005$) prefer CBTL after giving feedback about the first meal. (d) CBTL predicts user responses with increasing accuracy over time.}
    \label{fig:web-study-results}
    \vspace{-1.5em}
\end{figure}

\textbf{Study Design. }
We consider the assisted feeding scenario introduced in Section~\ref{sec:intro}.
Through a custom website (see Appendix~\ref{app:web-study}), \numparticipants{} participants are presented with a sequence of five meals:
\begin{tightlist}
    \item{\emph{Meal 1:}} \underline{Alone} at a \underline{rectangular} table with \underline{french fries} and \underline{ketchup and BBQ sauce}.
    \item{\emph{Meal 2:}} \underline{With friend} at a \underline{circular} table with \underline{raw vegetables} and \underline{ranch dressing and hummus}.
    \item{\emph{Meal 3:}} \underline{With TV} at a \underline{circular} table with \underline{potato wedges} and \underline{ketchup and BBQ sauce}.
    \item{\emph{Meal 4:}} \underline{Alone} at a \underline{circular} table with \underline{carrot sticks} and \underline{ranch dressing and hummus}.
    \item{\emph{Meal 5:}} \underline{With friend and TV} at a \underline{rectangular} table with \underline{tater tots} and \underline{ketchup and BBQ sauce}.
\end{tightlist}

For each meal, the robot must choose (1) the plate pose; (2) the above-plate utensil pose and robot joint positions; (3) the dip, if any; (4) the ready signal (button, auto-continue, or mouth open); and (5) whether to speak to the user verbally.
We compare CBTL and No Personalization, which is nontrivial here: it does not know if the user prefers ketchup or BBQ sauce for fries, but it does know that peanut butter would be unusual, given its LLM access.
Before each meal, the participant is presented with the predictions of CBTL and No Personalization (``Option A'' or ``Option B'', random per meal) and asked which set they prefer on a 5-point Likert scale.
The participant is then asked about their ideal dip, ready signal, and verbosity, as well as whether their view would be occluded in different egocentric images rendered from the CBTL-selected plate pose and robot joint positions.
CBTL learns from this feedback and adapts for subsequent meals.

\textbf{Results and Analysis. }
In Figure~\ref{fig:web-study-results}(c), we see that participants significantly prefer ($p < 0.005$, Wilcoxon-Signed Rank test) CBTL for every meal after the first.
In Figure~\ref{fig:web-study-results}(d), we also see that the choices made by CBTL increasingly align with the user's reported choices for each of the categorical options (dip, ready signal, verbosity).
Together, these results suggest that CBTL is able to rapidly and continually personalize to real humans.
See Appendix~\ref{app:web-study} for additional analysis.

\subsection{Real-Robot Demonstration}

\begin{wrapfigure}{r}{0.42\textwidth}
  \centering
  \vspace{-3.2em}
  \includegraphics[width=0.4\textwidth]{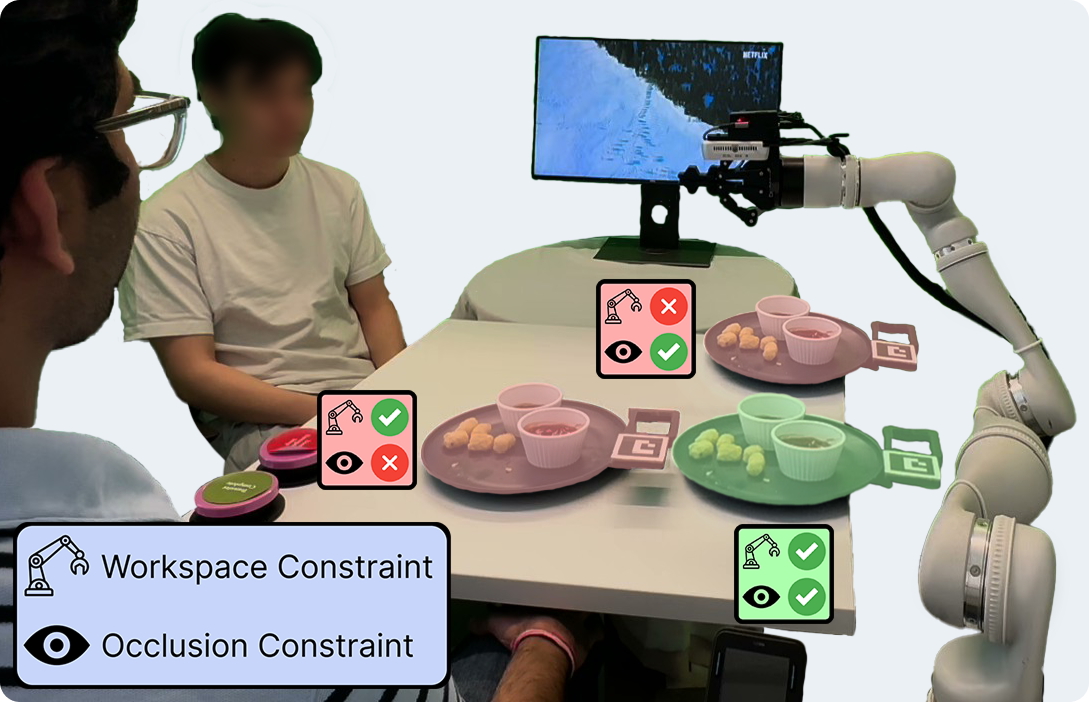}
  \caption{CBTL satisfies given workspace and personalized occlusion constraints.}
  \vspace{-1em}
  \label{fig:robot-csp}
\end{wrapfigure}

We next demonstrate CBTL on a real mealtime-assistance system (see Appendix~\ref{app:real-robot}).
A study participant completes Meals 1-4 through the web interface and then has Meal 5 in the real world (Figure~\ref{fig:front}).
The plate is initially placed in front.
Without any further feedback, CBTL optimizes the plate pose to avoid violating the user's occlusion preferences (Figure~\ref{fig:robot-csp}).
After serving a few bites, a drink (which was not in previous meals) is added to the table.
The robot reuses the personalized occlusion constraint from feeding and repositions the plate and the drink to avoid occlusions during drinking.
This experiment thus demonstrates not only the real-world applicability of CBTL, but also the ability of CBTL to rapidly generalize.

%% file: conclusion.tex
\textbf{Discussion.}
In safety-critical applications such as robot-assisted feeding, the extreme consequences of violating certain constraints makes it tempting to engineer systems that are entirely inflexible.
But users with diverse needs and preferences consistently voice the importance of personalization~\cite{jenamani2024flair,jenamani2025feast,nanavati2023design}.
As we continue to work towards robots that assist with not only feeding, but also cooking, cleaning, and countless other activities, we must develop task-independent personalization techniques that maintain flexibility without sacrificing safety or competency.
Coloring Between the Lines in CSP null spaces offers a unified and practical path toward addressing these challenges.

%% file: limitations.tex
\section{Limitations}
\label{sec:limitations}

One limitation of the current work is that we do not address how CSP generators should be systematically designed.
We took inspiration from PDDLStream~\cite{garrett2020pddlstream} to implement the generators used in our experiments, but we otherwise leave the question open.
Although this flexibility is a strength, it also places a substantial implementation burden on  robot developers to create their own CSP generators.
To partially mitigate this burden, one avenue for future work is automatically synthesizing personalized constraints in their entirety, rather than adapting pre-specified parameters.

Another related concern is that the extent to which safety and competency are guaranteed by CBTL depends on the implementation of the CSP generators.
This limitation is consistent with prior work on safety~\cite{wachi2024survey} and TAMP~\cite{garrett2021integrated}, which similarly assume that constraints are given and correct.
Nevertheless, further work is needed on methods that verify safety and competency.

Another limitation of the present work is that we restricted our focus to learning personalized \emph{constraints}, when certain preferences could be more naturally described as \emph{costs}.
For example, a user may wish to maximize the speed of the robot subject to safety constraints.
Our personalized constraints could be converted into costs, and we could use the same mathematical program that we use for active learning but with cumulative probability maximized instead of entropy.
However, non-probabilistic cost functions may be more appropriate in some settings, and it remains unclear how different costs should be combined when there are multiple within the same CSP.

One possible objection to the general CSP-based paradigm is that the robot should not be constantly planning and re-planning, especially when it is performing the same tasks repeatedly.
However, the extent to which the robot is ``planning'' is a function of the difficulty of the CSP.
A robot that generates and immediately solves ``easy'' CSPs would resemble a robot with a reactive policy.

%% file: acknowledgments.tex
\section{Acknowledgments}
This work was partly funded by National Science Foundation IIS \#2132846, and CAREER \#2238792.
We are grateful to Nishanth Kumar, Aidan Curtis, Yuki Wang, Jakob Thumm, and Chris Agia for feedback on the project, and to all members of the EmPRISE lab for maintaining computational and robot infrastructure.

%% file: appendix.tex
\section{CSP Details: Generating, Solving, and Learning}
\label{app:csp-details}

In this section, we describe details for the CSP generation, solving, and learning techniques that we use in our experiments.
Coloring Between the Lines is agnostic to these details.

\textbf{CSP Generation. }
Each environment has an associated CSP generator.
The CSP generator monitors all observations and actions and is sometimes queried to produce a new CSP.
When queried, the CSP generator outputs: (1) a CSP $(\cspvariables, \cspconstraints)$; (2) an initial candidate solution $\cspcandidatesolution_0$; (3) conditional samplers for the solver (see below); (4) a solution-conditioned policy $\policy(\action_t \mid \history_{t}, \cspsolution)$; and (5) a solution-conditioned termination condition $\termination(\history_{t}, \cspsolution)$.
The logic for CSP generation is environment-specific.
In the PyBullet environments (Cleaning and Books), we factorize CSP generation into \emph{skills}---for example, moving to a shelf, picking a book, moving to the human, and handing over the book.
The logic for sequencing skills is pre-specified in our experiments; this could be generalized using task planning~\cite{garrett2020pddlstream}.
Each skill contributes variables (e.g., $\mathrm{grasp} \in \text{SE(3)}$ for picking), constraints (e.g., $\mathrm{collisionFreeGrasp(grasp)}$), a subpolicy, a sub-termination condition, and zero or more samplers to the CSP generation process.
Skills are reused within and between CSPs.
For example, the movement skill uses a motion planner (BiRRT) to move between any two points in the environment.

\textbf{CSP Solving. }
We use a simple and flexible sampling-based approach to solve CSPs.
The solver performs a random walk over candidate solutions starting with the initialization $\cspcandidatesolution_0$ produced by the CSP generator.
Each subsequent candidate $\cspcandidatesolution_{i+1}$ is generated from a sampler $\cspsampler(\cdot \mid \observation, \cspcandidatesolution_i)$ where $\observation$ is the most recent observation at the time of CSP generation.
Each $\cspsampler$ typically modifies only a small subset of variable values.
At every step in the random walk, a sampler is randomly selected from the set of samplers generated by the CSP generator.
The first candidate that satisfies all constraints is returned by the CSP solver.
We set large timeouts for the CSP solver and do not hit the limit in any of our experiments, i.e., CSP solving always succeeds.
For entropy-based active learning, where we have a mathematical program, we need to maximize an objective (mean entropy) in addition to satisfying constraints.
We use the same sampling-based approach, but rather than terminating when the first solution is found, we continue until we find $N_{\text{improve}}=500$ solutions with objective values greater than or equal to the best seen value and then return the best found solution.

\textbf{CSP Supervised Constraint Learning. }
We update personalized constraint parameters with supervised learning and LLMs.
We provide details for supervised learning here and for LLMs in the section below.
The following personalized constraints use supervised learning in our experiments: (1) $\mathrm{userEnjoysMeal}$ in Cooking; (2) $\mathrm{surfacePreferred}$ in Cleaning; (3) $\mathrm{poseReachable}$ in Books; and (4) $\mathrm{occlusionFree}$ in the real-robot mealtime assistance system.
The $\mathrm{surfacePreferred}$ model simply records, for each surface where cleaning was attempted, whether or not the user admonished the robot on the last attempt; any surface not yet cleaned is assumed preferred.
The three other constraints each use one or more instantiations of a model $\cspconstraint(\cspvalue) = 1[\theta_0 \le f(\cspvalue) \le \theta_1]$ where $\theta_0, \theta_1 \in \mathbb{R}$ and $f(\cspvalue) \in \mathbb{R}$ derives a 1D value from the CSP variable candidate.
For example, the $\mathrm{poseReachable}$ constraint in the Books environment computes the Euclidean distance between the human's resting hand position and the candidate handover.
We analytically compute a posterior distribution over $\theta_0$ and $\theta_1$ given positive and negative examples of the constraint.
The maximum likelihood values are used for the constraints and the full posterior is used when computing entropy in active learning.

\section{Large Language Model Constraint Details}
\label{app:llm-details}

We now provide details for the LLM-based personalized constraints used in our experiments.
These include: (1) $\mathrm{userEnjoysBook}$ in Books; and in the real robot experiments, (2) $\mathrm{dipPreferred}$; (3) $\mathrm{signalPreferred}$; and (4) $\mathrm{verbosityPreferred}$.
As described in the main text, each constraint is characterized by a $\textsc{ConstraintPrompt}$ and a $\textsc{LearningPrompt}$.
We use \texttt{gpt-4o-mini} through the OpenAI API for all LLM queries~\cite{openai2024gpt4omini}.

For entropy-based active learning, we must predict the \emph{probability} that a constraint is satisfied.
Consistent with other work, we find that the LLM next-token probabilities  do not provide a useful measure of uncertainty; the LLM typically predicts values with extremely high probability, even in cases where its predictions are incorrect.
The following trick mitigates this issue in our experiments.
Rather than asking the LLM to return True or False, we ask it to return an answer on a scale from 0 to 10.
For example, for $\mathrm{userEnjoysBook}$, we ask: ``How much would the user enjoy the book on a scale from 0 to 10, where 0 means hate and 10 means love?''
We interpret a response of 8 out of 10 to mean that if the user was asked 10 times whether or not they enjoy a book, they would say yes 8 times and no 2 times.
Under this interpretation, the expected probability is $\sum_{i=0}^{10} \frac{i}{10} p_i$, where $p_i$ is the LLM's predicted probability for the next token $i \in \{0, 1, \dots, 10\}$.
The full prompt is below.

\vspace{-1em}
\begin{emptybox}[\textnormal{\textsc{ConstraintPrompt} for} $\mathrm{userEnjoysBook(book, \constraintparameters)}$]
Book description: $\mathrm{book}$. User description: $\theta$. I will ask you to rate how much the user would enjoy this book on a scale from 0 to 10. IMPORTANT: if you have already seen this book before, please answer based on the user's previous enjoyment of the book. How much would the user enjoy the book on a scale from 0 to 10, where 0 means hate and 10 means love? Return a number from 0 to 10 and nothing else.
\end{emptybox}
\vspace{-1em}

The role of the \textsc{LearningPrompt} is to condense the relevant experience collected thus far into a natural language summary, which serves as the constraint parameter $\constraintparameters$.
In our experiments, it was feasible to provide all feedback to the LLM and ask it to summarize; in general, it may be better to provide only the most recent feedback and ask for an update to the most recent parameter, i.e., learn incrementally.
To maintain uncertainty in the face of limited data, we found it useful to ask the LLM to generate multiple possible explanations.
The entire response is used as $\constraintparameters$.
The prompt for the Books environment is shown below.

\vspace{-1em}
\begin{emptybox}[\textnormal{\textsc{LearningPrompt} for} $\mathrm{userEnjoysBook(book, \constraintparameters)}$]
You are a helpful assistant that infers a user's book preferences based on their feedback. Your current estimate of their preferences is $\constraintparameters$. You have received the following feedback from the user: $\langle$all feedback in $\dataset \rangle$. Based on this history, provide an updated description of the user's book preferences. Your description should be in the following format: ``I know that the user likes the following books: [list of books]  and they do not like the following books: [list of books]. Based on this, here are some possible summaries of their preferences: 1. [summary of preferences] 2. [summary of preferences] 3. [summary of preferences].'' Return this description and nothing else. Do not explain anything.
\end{emptybox}

\section{Additional Results}
\label{app:additional-results}

We now present additional simulation experiments and results.

\textbf{Unsatisfiable Personalized Constraints.}
An additional advantage of CBTL is that there is natural recourse when personalized constraints are unsatisfiable: re-planning without personalized constraints.
For example, in the Cooking environment, suppose that ingredient quantities are limited such that the user's preferences cannot be fully satisfied, but a meal can still be made.
Given the explicit control flow of CBTL, the robot can notify the user that their preferences cannot be satisfied and give them the option of a non-personalized solution (some meal).

\begin{wraptable}{r}{0.55\textwidth}  
  \setlength{\intextsep}{0pt}          
  \centering
  \vspace{-1.5em}
  \begin{tabular}{@{}lrr@{}}           
    \toprule
    Method & Task Success & Preferences Met\\
    \midrule
    CBTL (Replan) & 1.00 $\pm$ 0.00 & 0.21 $\pm$ 0.09\\
    CBTL (No Replan) & 0.31 $\pm$ 0.11 & 0.95 $\pm$ 0.09 \\
    No Personalization & 1.00 $\pm$ 0.00 & 0.00 $\pm$ 0.00\\
    \bottomrule
  \end{tabular}
  \captionsetup{font=small}
  \caption{Planning with unsatisfiable personalized constraints in the Cooking environment (mean $\pm$ stdev).}
  \label{tab:unsatisfiable}
  \vspace{-1.5em}
\end{wraptable}

To illustrate this flexibility, we run an additional experiment in the Cooking environment where ingredient quantities are limited so that meals are always possible, but when each ingredient quantity respawns, there is only a $50\%$ chance that there will be enough to satisfy the user's preferences.
We compare CBTL with and without non-personalized re-planning to No Personalization across 10 seeds with 10 evaluation tasks per seed after 2500 training steps.
In Table \ref{tab:unsatisfiable}, we show that CBTL maintains perfect task success rates (meals served) by re-planning without personalized constraints after planning in the full CSP fails.
In contrast, CBTL (without replanning) consistently satisfies user preferences, but at the expense of task success; No Personalization never satisfies user preferences.
In future work, we plan to explore more sophisticated fall-back strategies (cf. minimum constraint removal~\cite{hauser2014minimum}) and  questions related to human-robot interaction with unsatisfiable preferences.

\textbf{Nonstationary Preferences.}
User preferences and functionalities can change over very long-term deployments.
In this additional experiment, we consider three variations of CBTL and assess the extent to which each can handle these nonstationary distributions.
In particular, we consider: (1) No Adapt, which assumes stationary user preferences; (2) User Reset, which resets the memory buffer in sync with preference shifts (analogous to explicit user feedback triggering relearning); and (3) Periodic Reset, which resets at fixed intervals (every $1000$ steps) independent of preference shifts.
We compare these approaches in a variation of the Cooking environment where hidden user preferences about ingredient quantity and temperature are periodically perturbed.
Concretely, every $3000$ simulation steps, we shift the user's preference intervals for each ingredient quantity and temperature by a scalar sampled uniformly at random from $[0.9, 1.1]$.

\begin{wrapfigure}{r}{0.5\textwidth}
  \centering
  \vspace{-2em}
  \includegraphics[width=0.5\textwidth]{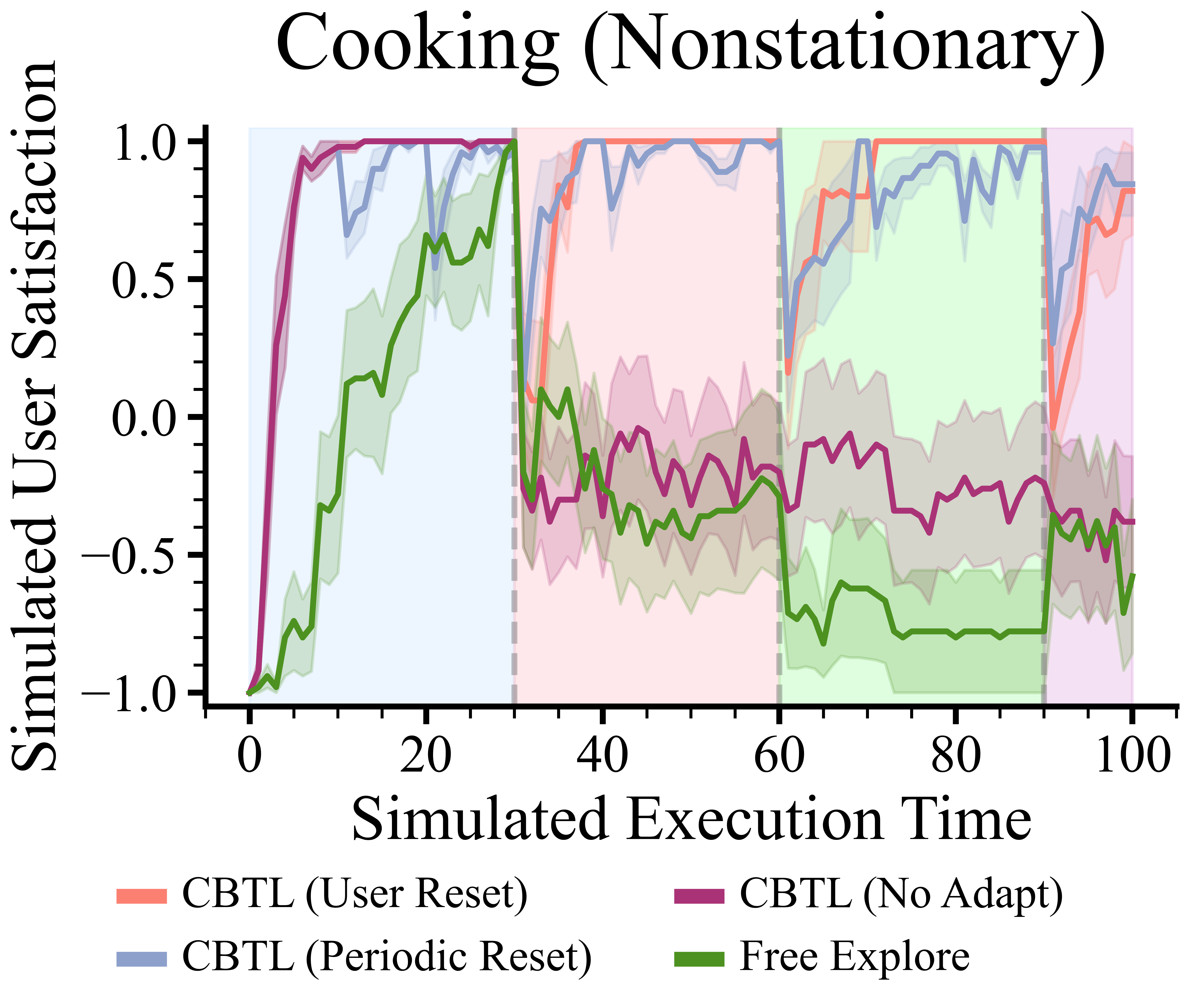}
  \captionsetup{font=small}
  \caption{CBTL adapts to shifting user preferences in nonstationary environments.\protect\footnotemark}
  \label{fig:nonstationarity}
  \vspace{-2.5em}
\end{wrapfigure}
\footnotetext{CBTL (User Reset) is identical to CBTL (No Adapt) before the first preference shift ($3000$ steps).}

As shown in Figure~\ref{fig:nonstationarity}, CBTL (User Reset) rapidly relearns personalization constraints after each shift, achieving consistent high satisfaction with minimal degradation during the relearning process. CBTL (Periodic Reset) similarly recovers quickly though with more instability due to unnecessary resets. In contrast, CBTL (No Adapt) and Free Explore fails to adapt and suffers from degrading performance after each shift. These findings highlight the flexibility of extending CBTL to nonstationary settings and motivate future work on automatically detecting preference shifts for even more adaptive lifelong personalization.

\section{Simulation Experiment Details}
\label{app:sim-experiment-details}

In this section, we provide additional details about the experimental setup, environments, and approaches for the simulation results presented in the main paper.

\textbf{Experimental Setup. }
All approaches and all environments are run over 10 random seeds.
We use a CPU-only cluster where machines with 64 GB memory run Ubuntu 20.04.
Each (approach, environment, seed) trial has a timeout of 6 hours, but typically takes between 30 and 90 minutes.

\textbf{Environments. }
We now provide further details about the simulation environments to supplement the descriptions given in the main text.

\emph{Cooking: }
We use two ingredients and four pots in the main experiments.
Ingredients heat up at a fixed rate when in pots and respawn after meals are served.
The following actions are available in the environment: (1) add a certain quantity of a certain ingredient to a certain pot; (2) move a certain pot to the stove at a certain position; (3) wait; (4) serve meal.
When the meal is served, the user reports whether or not they like the temperature and quantity of each ingredient that is currently in some pot based on hidden preference specifications.
A preference specification for an ingredient temperature or quantity is characterized by an upper and lower bound.
User satisfaction is $-1$ if any preference is violated when a meal is served and $1$ otherwise.
We use 10 tasks to evaluate the approach checkpoints, where each task finishes when some meal is served.
Evaluation is performed after every $100$ actions and trials are run for a total of $10,000$ actions.

\emph{Cleaning: } The robot must pick up and use a dusting tool to wipe surfaces (see video).
There are seven surfaces in the environment (three shelves in a bookcase and four side tables), which are spread out around the environment, necessitating robot base movement.
Dust respawns quickly after the robot cleans.
The user has hidden preferences that only two surfaces should be cleaned by the robot.
If the robot starts to clean one of the other surfaces, the user expresses displeasure.
Actions in the environment are delta joint positions for the robot arm, delta positions and rotations for the robot base, and open/close for the robot gripper.
Skills implemented as part of the CSP include base movement, arm movement, picking, placing, and wiping.
The movement skills use BiRRT for motion planning with IKFast for inverse kinematics and PyBullet for collision checking.
Picking and placing use simple grasp and placement samplers that assume cuboid objects.
Wiping uses motion planning to get into an initial position before executing a predefined up-down, left-right motion.
We use 10 tasks to evaluate approach checkpoints.
For each task, we randomly sample one surface to cover with obstacles (preventing wiping).
The robot cleans one surface and then the user reports $-1$ or $1$ satisfaction according to their hidden preferences.
Evaluation is performed after every $500$ actions and trials are run for a total of $5,000$ actions.

\emph{Books: } The robot must hand over books to a user lying in bed.
In our main experiments, we use the following nine well-known books:
\begin{tightlist}
    \item \textit{Cosmos} by Carl Sagan.
    \item \textit{Pride and Prejudice} by Jane Austen.
    \item \textit{The Hitchhiker's Guide to the Galaxy} by Douglas Adams.
    \item \textit{The Immortal Life of Henrietta Lacks} by Rebecca Skloot.
    \item \textit{The Diary of Anne Frank} by Anne Frank.
    \item \textit{Into the Wild} by Jon Krakauer.
    \item \textit{Moby Dick} by Herman Melville.
    \item \textit{The Lord of the Rings} by J. R. R. Tolkien.
    \item \textit{And Then There Were None} by Agatha Christie.
\end{tightlist}

The user has the following hidden natural language preferences: ``I only like two kinds of books: fictional romances, and grand, nonfictional reflections on the universe.''
When they are handed a book, short feedback is generated using an LLM with access to the hidden preferences.
The user also has a hidden functional range of motion.
If the robot attempts to hand over a book outside their range of motion, the user responds: ``I can't reach there.''
Actions and skills are the same as in the Cleaning environment.
We use 10 task to evaluate approach checkpoints.
For each task, a random subset of the books are available.
The user reports $1$ satisfaction only if the robot hands over a book that they enjoy on the first attempt.
Evaluation is performed after every $500$ actions and trials are run for a total of $5,000$ actions (same as Cleaning).

\textbf{Baselines. }
We next provide additional details about the baselines used in simulation experiments.

\emph{No Personalization.} This baseline is the same as CBTL, except personalized constraints are never updated.
For example, in Books, the user preferences are initially ``unknown'', and they remain unknown in perpetuity.

\emph{Free Explore.} This baseline is the same as CBTL, except that whenever a CSP is generated during exploration (i.e., any time other than during the checkpoint-loaded evaluation tasks), personalized constraints are not included.
We call this ``Free Explore'' because the baseline is free to explore any solutions that satisfy the non-personalized constraints.

\emph{Exploit Only.} This baseline is the same as CBTL, except that whenever a CSP is generated during exploration, the personalized constraints (with their current learned parameters) are always included.
We call this ``Exploit Only'' because the baseline exploits its current learned models at all times.
Like in reinforcement learning, this approach can easily get stuck in local minima.

\emph{Epsilon-Greedy.} This baseline is the same as CBTL, except that whenever a CSP is generated, it includes personalized constraints (all or nothing) with probability $\epsilon=0.5$.
We varied the value of $\epsilon$ and found that $\epsilon=0.0$ and $\epsilon=1.0$ match Free Explore and Exploit Only respectively, but there is no clear middle ground that avoids the pitfalls of both.
We also note that Epsilon-Greedy and Exploit Only obtain equal performance in the Cooking results.
This is due to the fact that both get trapped in the same local minima and the fact that we carefully control random seeding throughout experiments.
Discrepancies emerge with lower values of $\epsilon$ and more random seeds.

\section{Web-Based User Study Details}
\label{app:web-study}

In this section, we provide additional details and analysis for our web-based study.

\textbf{Demographics. }
We recruited $\numparticipants{}$ participants who completed the study online.
The participants were not compensated.
They ranged in age from 19 to 68 years old with a mean of 31.
Given the options Male, Female, Non-binary, Prefer Not To Say, and Other, 36 participants selected Male and 24 selected Female.
We asked the participants if they had prior robot experience (29 yes, 31 no) and any experience being fed as an adult (13 yes, 47 no).

\textbf{Website Design. }
We built a custom website for the study using pure JavaScript.
Although the study presents to the participant as a typical survey, the participant's earlier answers impact the choices that they are later shown, which makes it difficult to use existing survey platforms.
All together, there are 228,637,728 ``paths'' that a participant can take.
However, we can factorize over question type (occlusion, verbosity, ready signal, dip preferences).
This makes it possible to pregenerate all possible model predictions offline.
We did not record time, but participants reported that the survey took approximately 15 minutes.

\begin{figure}
    \centering
    \includegraphics[width=\linewidth]{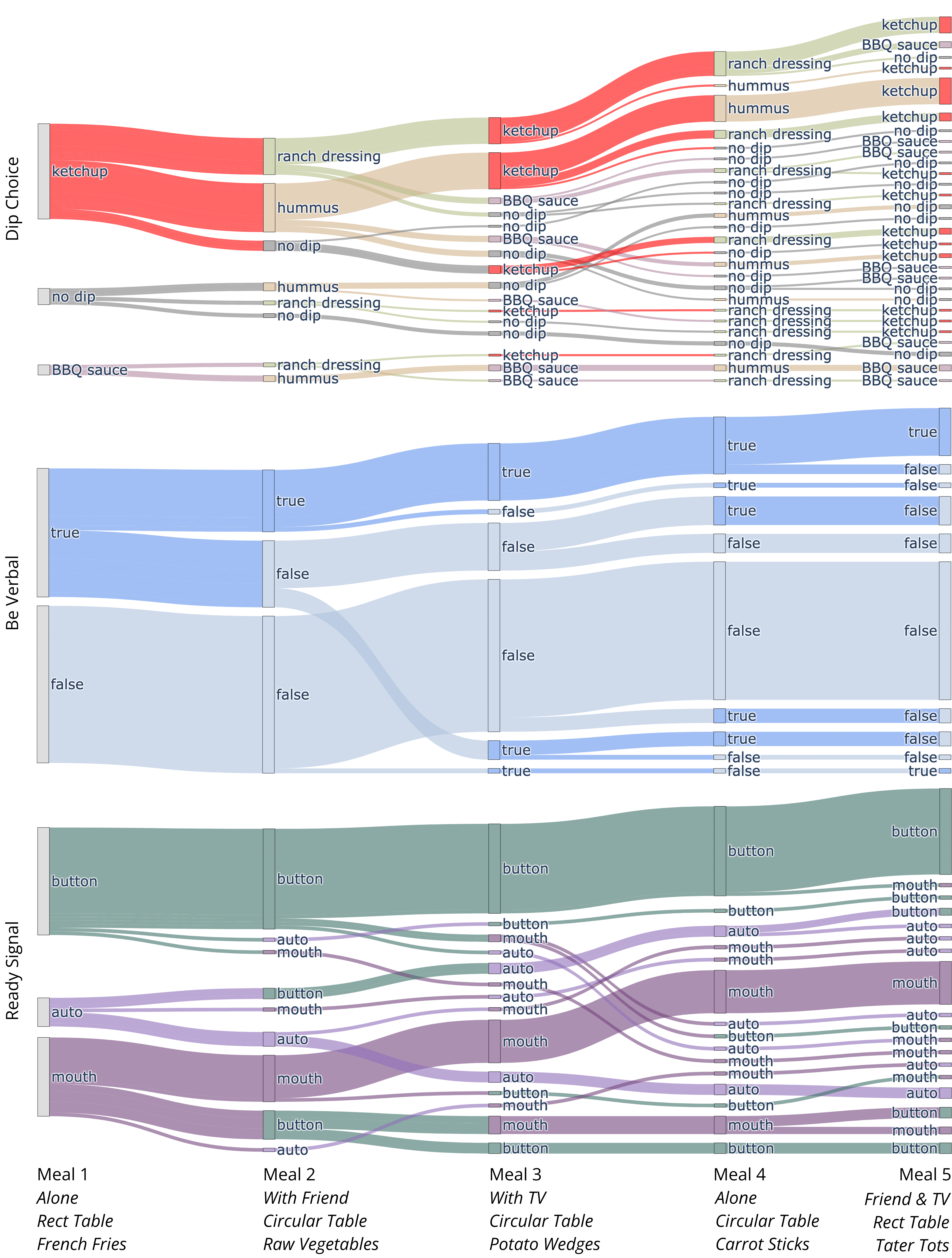}
    \caption{Sankey Diagrams for the three categorical questions in the web study. We see diversity in all three categories. This confirms that the predictive ability of CBTL is indicative of personalization.}
    \label{fig:web-study-sankey}
\end{figure}

\textbf{Additional Analysis. }
In the main paper, we presented results that show (1) participants prefer CBTL after the first meal; and (2) CBTL is increasingly able to predict participant responses over time.
To further confirm CBTL's personalization capabilities, we present results in Figure~\ref{fig:web-study-sankey} showing that there is substantial variation between participants in terms of their preferences.
For example, in the case of dip choice, there were 29 distinct paths for 60 participants.
This diversity underscores the importance of personalization and confirms that CBTL personalizes to individual users.

Returning to the main results, in Figure~\ref{fig:web-study-results}(c), we note the absence of a monotonic improvement in terms of preference for CBTL.
For example, participants had a stronger preference for CBTL on Meal 3 than on Meal 4 on average.
We hypothesize that this is primarily due to the fixed meal ordering and the fact that participants may have varying overall preference strengths for different meals.
For instance, a participant may strongly prefer ketchup in Meals 1, 3, and 5, and weakly prefer hummus in Meals 2 and 4.
This hypothesis is supported by the fact that CBTL's ability to predict participant responses \emph{does} monotonically improve (Figure~\ref{fig:web-study-results}(d)).

\section{Real Robot Details}
\label{app:real-robot}

For our real robot demonstration, we use a 7-DoF Kinova Gen3 arm~\cite{kinova}.
The arm is equipped with an Intel RealSense RGB-D in-hand camera~\cite{camera} and a modified Robotiq 2f-85 gripper~\cite{robotiq} with custom fingers.
To assist with feeding, the robot can pick up a fork-tipped feeding utensil from a tool mount.
The utensil includes two onboard motors that facilitate scooping and twirling motions, enabling the robot to dexterously pick up food items from a plate and bring them to the user’s mouth. The robot can also pick up a mug leveraging a custom handle with an attached AR tag, which is affixed to the mug, and bring it to the user for sipping. Similarly, it can reposition the plate, which is attached to a matching custom handle.